\documentclass[10pt,twocolumn,letterpaper]{article}

\usepackage[pagenumbers]{cvpr} 

\usepackage{graphicx}
\usepackage{amsmath}
\usepackage{amssymb}
\usepackage{booktabs}
\usepackage{pifont}
\usepackage{algorithm}
\usepackage{algorithmic}
\usepackage{multirow}
\usepackage{url}

\usepackage[pagebackref,breaklinks,colorlinks]{hyperref}

\usepackage[capitalize]{cleveref}
\crefname{section}{Sec.}{Secs.}
\Crefname{section}{Section}{Sections}
\Crefname{table}{Table}{Tables}
\crefname{table}{Tab.}{Tabs.}

\def\cvprPaperID{4378} 
\def\confName{CVPR}
\def\confYear{2023}

\begin{document}

\title{Addressing Data Heterogeneity in Decentralized Learning via Topological Pre-processing}

\author{Waqwoya Abebe, Ali Jannesari\\
Department of Computer Science\\
Iowa State University
\\
{\tt\small \{wmabebe, jannesar\}@iastate.edu}
}
\maketitle

\begin{abstract}
Recently, local peer topology has been shown to influence the overall convergence of decentralized learning (DL) graphs in the presence of data heterogeneity. In this paper, we demonstrate the advantages of constructing a proxy-based locally heterogeneous DL topology to enhance convergence and maintain data privacy. In particular, we propose a novel peer clumping strategy to efficiently cluster peers before arranging them in a final training graph. By showing how locally heterogeneous graphs outperform locally homogeneous graphs of similar size and from the same global data distribution, we present a strong case for topological pre-processing. Moreover, we demonstrate the scalability of our approach by showing how the proposed topological pre-processing overhead remains small in large graphs while the performance gains get even more pronounced. Furthermore, we show the robustness of our approach in the presence of network partitions. 
\end{abstract}

\section{Introduction}

Advances in distributed machine learning have brought forth the rise of Federated Learning (FL) \cite{mcmahan2017fedavg, yu2021adaptive} and Decentralized Learning (DL) \cite{colin2016gossip, pmlr-v97-assran19a}. While FL aims to train a centralized model using decentralized data, DL attempts to train decentralized models using decentralized data. In particular, DL involves the participation of decentralized peer-to-peer (p2p) nodes that collaboratively train their local models in a manner where all local models ideally converge on the global data distribution \cite{kairouz2021advances}. A major issue in FL/DL environments is the inherent presence of non-Independent and Identically Distributed (non-IID) \cite{mcmahan2017fedavg} global data, commonly referred to as data heterogeneity.

Although DL applications have the natural advantage of scalability \cite{assran2019stochastic} for lack of a bottleneck like the FL global aggregation server \cite{lian2017can}, the absence of a central server and global model make it hard to mitigate data heterogeneity. For instance, with the knowledge of global gradient (i.e. by averaging all local gradients), FL servers are ideally positioned for computing the global gradient direction and relative local gradient drifts \cite{ozfatura2021fedadc} providing them a better means of addressing data heterogeneity. Replicating drift computation in DL requires knowledge of all local gradients by all nodes. I.e. a fully connected p2p network (complete graph of peers). Despite the optimal convergence of a fully connected DL graph \cite{bars2022yes} however, such a dense topology is infeasible as it incurs a large communication overhead especially as the system scales.

Intuitively speaking, peers in a linear DL topology are likely to converge slower than in a denser graph due to delays in parameter updates. As such, most prior work involving peer topology and performance has mainly focused on the spectral gap of the graph \cite{bars2022yes}. Recently however, the arrangement of nodes in local neighborhoods/sub-graphs have been shown to influence global convergence even among graphs of similar size and topology.  Works such as \cite{bellet2021d, bars2022yes} show how regional/neighborhood topology plays a pivotal role in influencing global convergence.  Similarly, \cite{dandi2022data} shows how the mixing weights of the graph can improve convergence. More specifically, these works show that constructing a peer topology, in a manner where local neighborhoods (sub-graphs) contain heterogeneous underlying data distribution (i.e. representative of the global distribution), can improve global convergence.

In light of this, we propose a scheme of constructing locally heterogeneous DL graphs. But unlike previous works that measure heterogeneity in terms of underlying data distribution, we measure heterogeneity in terms of model data representations (knowledge). Furthermore, instead of using model parameters to compute the divergence among nodes, we propose using a lightweight model substitute which we refer to as ``model proxy". Proxies serve as summaries that abstract a model's acquired knowledge. In particular, we use soft-labels outputted by a model on a global dataset as it's knowledge proxy. Given two proxies, we compute their similarity using the Kullback-Leibler (KL) divergence. This approach has a significant advantage in terms of maintaining peer data privacy since peers need not reveal their data distributions for ideal placement in the DL graph. Additionally, proxies have a considerably smaller footprint compared to their corresponding models and gradients. This makes proxies ideal for reducing communication overhead.

\begin{figure*}
    \centering
    {{\includegraphics[width=17cm]{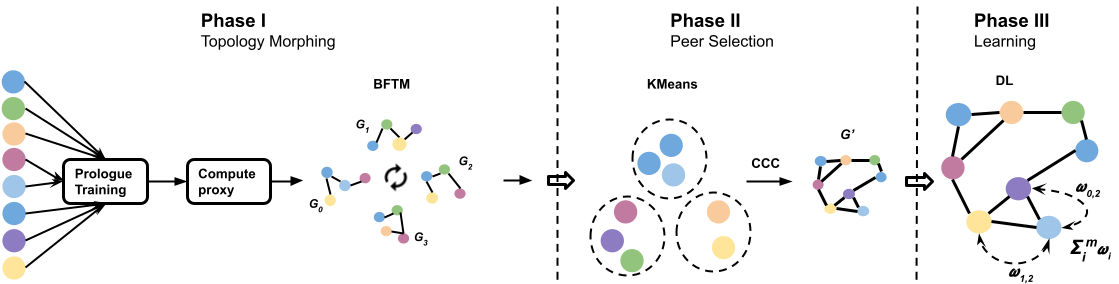} }}%
    \caption{System overview: {\large\textcircled{\raisebox{-0.9pt}{\small I}}} Conduct prologue training, compute proxy then iterate through a few rounds of BFTM until the similarity matrix is complete. {\large\textcircled{\raisebox{-0.9pt}{\small II}}} Cluster nodes into $M$ groups and construct a locally heterogeneous final graph $G'$. {\large\textcircled{\raisebox{-0.9pt}{\small III}}} Commence the DL process using a local FedAvg aggregation algorithm.}%
    \label{fig:overview}%
\end{figure*}

However, even simply computing pairwise proxy divergence is an expensive task. It requires broadcasting all proxies costing $O(N^2)$ overall downloads. To minimize this cost, we crafted a novel gossip algorithm involving only 1-hop communications and iterative graph morphing called Breadth-First-Topology-Morphing (BFTM). BFTM uses iterative graph restructuring and caching to speedup the computation of pairwise proxy similarities. It guarantees complete pairwise computation in $\Theta({\frac{\sqrt N}{\log N}})$ iterations and $\Theta(N\sqrt N)$ overall downloads, making it quick and communication efficient.

After completely populating a pairwise proxy similarity matrix, we determine constructing an optimally heterogeneous final graph is an NP-hard problem. Hence, we couple KMeans clustering with a simple Cross-Cluster Cliquing (CCC) algorithm as a heuristic to construct a locally heterogeneous final graph $G'$. Finally, we show how locally heterogeneous graphs outperform locally homogeneous graphs of similar size under various non-IID settings. Furthermore, we show how our approach is scalable and robust in the presence of network partitions. 

In summary, our contributions are: 
\begin{itemize}
\item Propose proxy-based topology construction for maintaining client data privacy and reducing communication overhead.
\item Develop a novel gossip algorithm (BFTM) to speed up proxy pair similarity computation.
\item Demonstrate that our method improves global convergence under various non-IID settings.
\item Show that the extra communication overhead our approach introduces remains small in large graphs, while the performance gains get more pronounced. 
\item Show that our method is robust under network partitions.
\end{itemize}

This paper is organized as follows: section II highlights Related Works, section III discusses the proposed Methodology, section IV showcases our Evaluation, and section V presents the Conclusion.

\section{Related Works}

Two key challenges slow down model convergence in DL. These are slow spread of information resulting in weight update delays and local data heterogeneity \cite{dandi2022data}. Several weight aggregation schemes have been proposed to tackle data heterogeneity in the FL setting. These include FedProx \cite{li2020federated}, FedNova \cite{wang2020tackling}, SCAFFOLD \cite{karimireddy2020scaffold}. Likewise, several algorithmic solutions have been proposed to tackle data heterogeneity under the DL setting \cite{tang2018d, nedic2017achieving, koloskova2021improved}.

On the other hand, peer topology has been a topic of interest in several DL studies. Despite the limited bandwidth usage of sparse topologies, \cite{pmlr-v108-neglia20a} showed how peer sparsity can speed-up wallclock time convergence since it intrinsically mitigates the straggler problem. \cite{9481923} proposed a means of jointly improving energy efficiency and convergence by constructing optimal topology.  

Few works consider the effect of DL topology in terms of improving performance with regards to data heterogeneity. In RelaySum \cite{vogels2021relaysum}, authors propose an aggregation mechanism over the DL spanning tree to reduce the effect of attenuating weight aggregation. This procedure helps improve the effect of non-IID imbalance that is worsened by 1-hop communication delays that cause weakened (attenuated) weight updates. In \cite{bellet2021d}, authors presented a locally heterogeneous D-clique p2p network topology to mitigate the non-IID problem. They constructed their DL network with a set of D-cliques in a manner where each clique possesses underlying data distribution that approximates the global data distribution. They explore several final graph topologies such as a ring of cliques, and a clique of cliques. In \cite{bars2022yes}, authors proposed how optimizing neighborhood heterogeneity improves global convergence among DL peers. In particular, they measure neighborhood heterogeneity as the difference between gradients of a local sub-graph and the whole graph. Furthermore, they propose optimizing neighborhood heterogeneity via the Frank-Wolfe algorithm with the ability to track the quality of the learned topology. Similarly, \cite{dandi2022data} propose a time-varying and data-aware mixing matrix to minimize local gradient drift. Particularly, they demonstrate how model convergence depends on the relationship between data heterogeneity and the mixing weights of the DL graph. 

In line with these works, we propose the construction of locally heterogeneous topology to mitigate the effects of data heterogeneity. However, we measure heterogeneity in terms of proxy similarities rather than underlying data distribution or gradients. This greatly reduces the cost of communication thus minimizing bandwidth. We also refrain from freely assuming knowledge of global information such as the global gradient, as computing such values requires gradient aggregation making it unscalable in practice. In stead, we rely on cheap local communication and graph restructuring to compute a global similarity matrix. Our approach preserves local data privacy since nodes need not reveal their data distributions for ideal placement in the final DL graph $G'$. Moreover, unlike previous works that focus on label-skew and single partitions, we consider various non-IID cases and the effect of network partitions on the training process.

\section{Methodology}

In this paper, we use the terms ``node" and ``peer" interchangeably. The overall process is divided into three phases; while phases I \& II serve as topological pre-processing, phase III involves the training process (Fig. \ref{fig:overview}). Phase I (Topology Morphing) starts with constructing a random undirected communication graph $G$ of size $N$ and degree  $M$. Nodes in $G$ represent the peers that participate in the DL training. In phase I, all nodes perform a one-time local training for a limited number of epochs after which they compute their proxy values. Following proxy computation, $G$ repeatedly restructures itself via the BFTM algorithm in order to compute pairwise proxy similarity with minimal data transfer. Phase I only lasts for a limited number of rounds until a global proxy similarity matrix is fully populated. Here, since there is no global server, a copy of the global similarity matrix is maintained by all nodes and updated in each BFTM iteration. In phase II (Peer Selection), equipped with the global similarity matrix, nodes have sufficient information to cluster peers into $M$ categories. After clustering, we apply a simple algorithm, which we call CCC, to construct a locally heterogeneous final graph $G'$. In phase III (Learning), peers commence DL by aggregating their neighborhood model parameters. 



\begin{figure*}
\centering
\begin{minipage}[t]{1\columnwidth}
\centering
  \includegraphics[width=10cm]{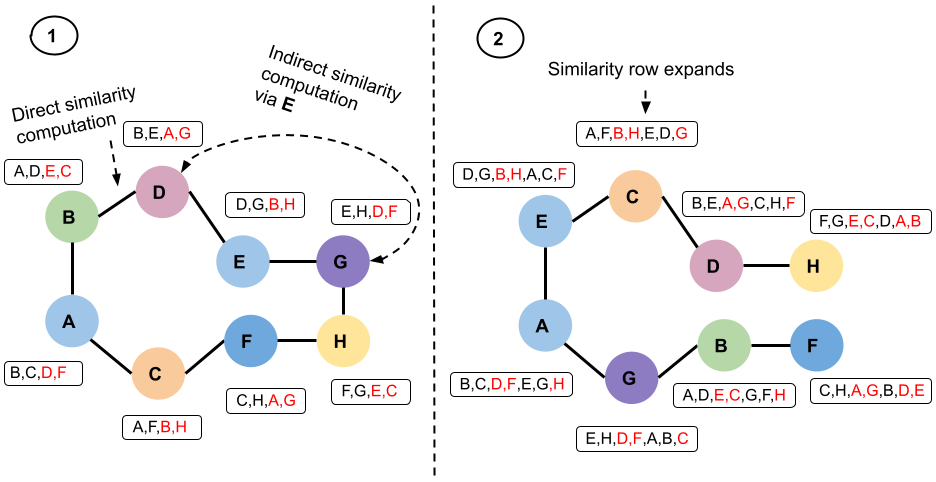}
\end{minipage}\hfill 
\begin{minipage}[t]{1\columnwidth}
\centering
  \includegraphics[width=5cm]{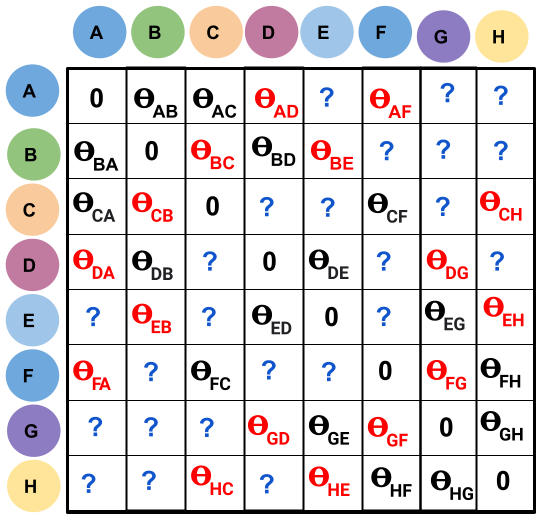}
  \end{minipage}
\caption{BFTM algorithm morphing an initial graph and populating the similarity matrix. Black similarity values are computed directly by downloading peer proxy. Red values are computed indirectly, through an intermediate neighbor. The similarity matrix shows the values computed at the end of step \textcircled{\raisebox{-0.9pt}{1}}.}
\label{fig:BFTM}%
\end{figure*}


\subsection{Phase I (Topology Morphing)}

\begin{figure}
\begin{minipage}[t]{1.0\columnwidth}
  \includegraphics[width=\linewidth]{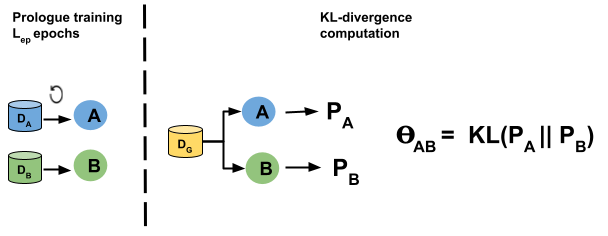}
\end{minipage}\hfill 
\caption{Computing proxy similarity using the KL-divergence between two proxies $P_A$ and $P_B$ (soft-labels outputted on a global dataset).}
\label{fig:proxy}%
\end{figure}

In this phase, we weave a special gossip protocol into a dynamically morphing graph. Topology morphing is an iterative process that attempts to compute pairwise similarities between all pairs of peers. Before morphing begins, nodes are required to engage in a one-time local training via Stochastic gradient Descent (SGD) for $L_{ep}$ epochs. We refer to this step as ``prologue training" since nodes resume local training in phase III. After prologue training, nodes compute their proxy value to abstract their model's locally acquired knowledge. We use soft labels outputted by individual models on a global dataset (Fig. \ref{fig:proxy}) to serve as the models' ``knowledge proxy". In this case, the global dataset is a limited and globally available dataset from a different distribution. Proxies concisely represent the knowledge learned by corresponding models. Nodes then compute pairwise peer similarities by iteratively exchanging and comparing these cheap and representative model proxies.

Before morphing begins, we assume that nodes know the identities/addresses of all other nodes. In practice this can be implemented via an announcement protocol whenever a node joins the DL network. In the morphing process, we start by constructing a random undirected graph of peers as follows. $G=\langle N,E \rangle \: $, where $ \: N$ is the set of peers and $E$ is a set of edges between peers. We initialize $G$ with a small degree $M$ to create a sparse graph. We deliberately minimize $M$, i.e. $M \ll N$ to reduce the system's communication overhead. Specifically, we set $M = \log N$. We subscript $G$ as $G_i$ (for the $i^{th}$ morphing round) to denote $G$'s various intermediate topologies (morphs) constructed during BFTM execution. At a high level, BFTM conducts five major tasks, the details are specified in Algorithm \ref{alg:BFTM}.

\begin{algorithm}[hbt!]
\caption{Breadth-First-Topology-Morphing}\label{alg:BFTM}
\label{alg:bftm}
\begin{algorithmic}
\REQUIRE{All nodes perform local training for $L_{ep}$ epochs}
\STATE $Nodes \leftarrow$ $\{n_0, n_1..., n_n\}$ 
\STATE $Queue \leftarrow \{\}$ 
\FOR{$n$ in $Nodes$}
\STATE $n$.cache() $\leftarrow \{\}$
\STATE $n$.missing() $\leftarrow Nodes$
\ENDFOR
\WHILE{$SIM\_MATRIX$ not full}
\STATE $Queue \leftarrow$ enqueue $n_0$ from $Nodes$
\STATE $G_{i+1} \leftarrow$ $\{\}$
\WHILE{$Queue$ not empty}
\STATE $n \leftarrow$ dequeue $Queue$[0]
\STATE $n$.neighbors() $\leftarrow$ pick $\{n_1,...,n_m\}$ from $n$.missing()
\STATE $n$.cache() $\leftarrow$ $n$.neighbors()
\STATE $n$.missing() $\leftarrow Nodes - n$.cache()
\STATE $S^n_{i+1} \leftarrow$ $S^n_{i}$
\FOR{$n_x$ in $n$.neighbors()}
\STATE $G_{i+1} \leftarrow$ add edge $\langle n,n_x \rangle$
\STATE $S^n_{i+1} \leftarrow$ add Sim($\langle n,n_x \rangle$)
    \FOR{$p_i$ in $n$.cache()}
        \STATE compute $\theta_{n_x,p_i}$
        \STATE $S^n_{i+1} \leftarrow$ add $\langle n_x, p_i,\theta_{p_i,n_x} \rangle$
    \ENDFOR
    \FOR{$n_y$ in $n$.neighbors()}
        \STATE compute $\theta_{n_x,n_y}$
        \STATE $S^n_{i+1} \leftarrow$ add $\langle n_x, n_y, \theta_{n_x,n_y} \rangle$
    \ENDFOR
\STATE $n$.missing() $\leftarrow$ remove $n_x$ 
\STATE $n$.cache() $\leftarrow$ add $n_x$ 
\STATE $n_x$.missing() $\leftarrow$ remove $n$ 
\STATE $n_x$.cache() $\leftarrow$ add $n$ 
\STATE $Queue \leftarrow$ add $n_x$
\ENDFOR
\STATE broadcast $S^n_{i+1} - S^n_{i}$
\IF{$|SIM\_MATRIX[n]| == |Nodes|$}
\STATE $Nodes \leftarrow$ remove $n$
\ENDIF
\ENDWHILE
\STATE $SIM\_MATRIX \leftarrow$ add all broadcast tuples
\ENDWHILE
\end{algorithmic}
\end{algorithm}

\begin{enumerate}
    \item Compute proxy (Done once).
    \item Share/Receive 1-hop proxies.
    \item Compute pairwise proxy similarities.
    \item Broadcast similarities and update local similarity matrix.
    \item Morph graph to connect un-encountered peers. Repeat step 2.
\end{enumerate}


In each BFTM round, peers download proxies from their 1-hop neighbors and compute pairwise peer similarities (Fig. \ref{fig:BFTM} left). This operation requires $O(NM)$ communication. A key optimization objective in phase I is to minimize proxy downloads. In light of this, the global dataset size is deliberately limited to minimize the proxy size and making it cheap to communicate.


After downloading neighboring proxies, each node $v$ computes similarities between its own proxy and that of it's neighbors. Here, we compute the KL-divergence $KL(P_{v}||P_{w})$ between two proxies (predictions) $(P_v, P_w)$ as shown in Eq. \ref{neighborhood_kl}. The KL-divergence measures the distance between the probability outputs of the predictions. The smaller the divergence, the closer the predictions are alluding to a similar knowledge representation and underlying data distribution. Peers also cache downloaded proxies to further accelerate BFTM via dynamic programming. Since $v$ caches neighboring proxies downloaded in previous BFTM iterations, it can quickly compute inter-neighbor proxy similarities between previously unpaired peers in the current BFTM round (Eq. \ref{inter_neighbor_angle}). 


\begin{equation}
    \label{neighborhood_kl}
    \theta_{vw} = KL(P_{v}||P_{w}) = \sum_{i=1}^N \sigma(P_{v}(x_i)^T log(\frac{\sigma(P_{v}(x_i))}{\sigma(P_{w})})
\end{equation}

\begin{equation}
\label{inter_neighbor_angle}
  \{\theta_{wx} : \;\; \forall \; \{w,x\} \in N \;\; and \;\; \{ w,x \} \in N^{cache}_{v} \}
\end{equation}

Assume set $p^v_i$ contains all proxies node $v$ has collected during rounds $1,..., i$. i.e. these proxies include node $v$'s own proxy and proxies of all previous $1$-hop neighbors.  At the start of round $i+1$, $v$ first downloads proxies $p'$ from its new 1-hop neighbors. Afterwards, it computes similarities between missing proxy pairs in $p^v_i$ and $p'$, as well as proxy pairs within $p'$. Thus in every round, $v$ computes a set of previously unknown pairwise similarity tuples ($\langle x,y,\theta_{xy} \rangle$, where $\{x,y\} \in N$), and updates its similarity matrix as shown in Eq. \ref{matrix_update}. Finally, $v$ updates its proxy cache $p^v$ as in Eq. \ref{grow_cache}. 

\begin{align}
\label{matrix_update}
S^v_{i+1} : S^v_{i} \; \bigcup \; \{\langle x,y,\theta_{xy} \rangle \} \; \bigcup \; \{\langle y,y',\theta_{yy'} \rangle\},  \nonumber\\ 
\forall\; x \in p^v_i, \; \{y,y'\} \in p' 
\end{align} 

\begin{equation}
\label{grow_cache}
  p^v_{i+1} = p^v_i \bigcup p'
\end{equation}

\begin{figure*}
\centering
\includegraphics[width=\linewidth]{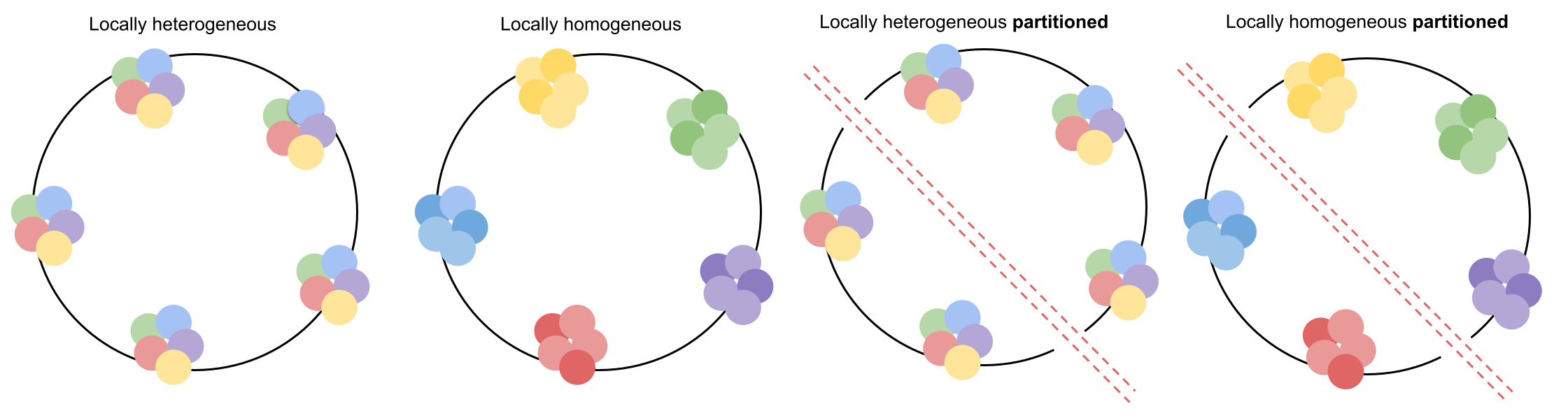}
 
\caption{Comparing locally heterogeneous and locally homogeneous final graphs with and without partitions.}
\label{fig:hetero-homo-ring}%
\end{figure*}

Nodes broadcast their newly added similarity tuples ($S^v_{i+1} - S^v_i$) at the end of each BFTM iteration. This communication overhead is significantly cheap since the similarity tuples are much smaller compared to the proxies. For instance, with similarity tuple size of $24$bytes, a $10,000$ node graph will broadcast $< 300$MB per round (about half the size of a single VGG11 model \cite{vgg}). 

Receiving nodes accordingly update their local similarity matrix. Similarity broadcast also informs nodes about indirect similarity computations made on their behalf. This indirect similarity computation allows nodes to narrow down their sets of previously un-encountered (missing) nodes from which they sample neighbors for the next round.

In every BFTM iteration $i$, each node's proxy will be paired with $\Theta(i^2M^2) = \Theta(i^2\log^2 N)$ other proxies. There is a maximum number of $N-1 \approx N$ proxies to be paired with. To determine how many iterations it takes to compute all proxy pairs, we equate $N = i^2\log^2 N$. This gives us $i = \Theta(\frac{\sqrt{N}}{log N})$. With node size of $2^{16} = 65536$ for instance, it only takes about $30$ BFTM rounds to completely populate the similarity matrix. Depending on the underlying data size however, learning might require a few hundred/thousand rounds. Thus, topology morphing rounds comprise of just a fraction of overall learning rounds. 

In cases where training data is limited (i.e. fewer learning rounds) or DL graph is large (i.e. more morphing rounds), we can apply a heuristic to make our approach feasible. For instance we can divide the nodes into smaller groups, or introduce early stopping in BFTM to produce partially complete similarity matrix.

\subsection{Phase II (Peer Selection)}

By phase II, nodes will have completely populated their local similarity matrix (Fig. \ref{fig:BFTM} right). The similarity matrix contains pairwise KL-divergence between all node pairs. Equipped with this information, it is now possible to construct a final graph by clumping nodes based on their proxy similarities.

Our objective is to construct a sparse and locally heterogeneous final graph $G'$. While sparsity increases communication efficiency, we show that local heterogeneity improves convergence. The key question here is, which nodes to pair with which ones in order to maximize convergence? As previous works \cite{bellet2021d, bars2022yes} show, ideal neighborhoods (sub-graphs) contain nodes possessing a relatively diverse/heterogeneous underlying data distribution. That's because the ideal neighborhood data distribution approximates the heterogeneous global data distribution. But rather than peeking into the nodes' underlying data distributions to construct the ideal topology, we propose constructing a topology based on the proxy similarity matrix. Our approach maintains node data privacy since we do not seek to know underlying data distributions. Furthermore, our approach has the potential to mitigate various flavors of non-IID conditions that could exist in practice.

Intuitively, therefore, we want to construct $G'$ such that all nodes are connected to neighbors in such a manner that the proxy divergence between a node and its neighbors is maximized. Simultaneously, we also want to maximize the proxy divergence amongst a node's neighbors. With this insight, we state each node $v$'s optimization objective for peer selection in Eq. \ref{optimization}:

\begin{align}
\label{optimization}
  argmax(\frac{1}{M}(\sum_{u}^{M}\theta_{vu})  + \sum_{\{u_i,u_j\}}^{M}\theta_{u_iu_j})) , \nonumber\\
  :  \{u,u_i,u_j\} \in M, \; \; \forall M \subset N
\end{align}

We can make a trivial reduction from the subset sum problem to show that the objective is NP-hard. The search space to find the optimal subset $M$ makes it impractical to apply naive search. To circumvent this issue, we applied unsupervised clustering as a heuristic and performed uniform heterogeneous sampling. In particular, we first feed the similarity matrix to a KMeans model to produce $M$ clusters (Fig. \ref{fig:overview}). This operation is reasonably fast (computes in seconds) for a large graph of $10$K nodes. If the matrix size exceeds the computation limit, a sub matrix can be extracted via uniform sampling as a substitute input. In either case, the KMeans model will group the peers into $M$ clusters. Next, we construct a locally heterogeneous final graph $G'$ by applying a CCC algorithm. CCC constructs a topology containing diverse cliques. More specifically, we experiment with a ring of cliques topology. i.e. CCC first constructs diverse cliques by uniformly sampling nodes from each cluster after which it chains the cliques to construct a ring topology as shown in Fig. \ref{fig:hetero-homo-ring}.

\subsection{Phase III (Learning)}

After constructing  $G'=\langle N',E' \rangle$, nodes can start engaging in the DL process. In each training round peers will aggregate their own weights alongside weights downloaded from their 1-hop neighbors. For instance, node $v$ aggregates its weights as shown in Eq. \ref{aggregate_weight}.

\begin{align}
\label{aggregate_weight}
  \omega_{i+1}^{v} = \frac{1}{M+1}(\omega_{i}^{v} + \sum_{u=1}^{m}\omega_{i}^{u}), \;\;  
   \forall \; \langle v,u \rangle \in E'
\end{align}

This step computes a local FedAvg \cite{mcmahan2017fedavg} algorithm in a decentralized manner as opposed to inside a central aggregating server. It is important to note that it takes $m$ rounds to receive updates from $m$-hop neighbors. Updates from distant peers will eventually ripple through the network albeit attenuating as they make their way through intermediate peers.

\section{Evaluation}

We conducted several experiments to test our proxy-based clustering hypothesis. It is important to note that in practice, it is not possible to determine before hand, the type and extent of non-IID imbalance that could plague a DL environment. As such, we can only simulate imbalance by introducing bias and noise into our dataset. To this end, \cite{li2022federated} provides a popular non-IID benchmark that broadly categorizes non-IID flavors into different sets. These include quantity-based label imbalance (\#label2), distribution-based label imbalance (labeldir), quantity skew, noise-based feature imbalance (feat-noise) and mixed non-IID settings. We conduct our experiments on four non-IID cases and show which scenarios best benefit from topological pre-processing. 

\begin{figure*}
    \centering
    {{\includegraphics[width=17cm]{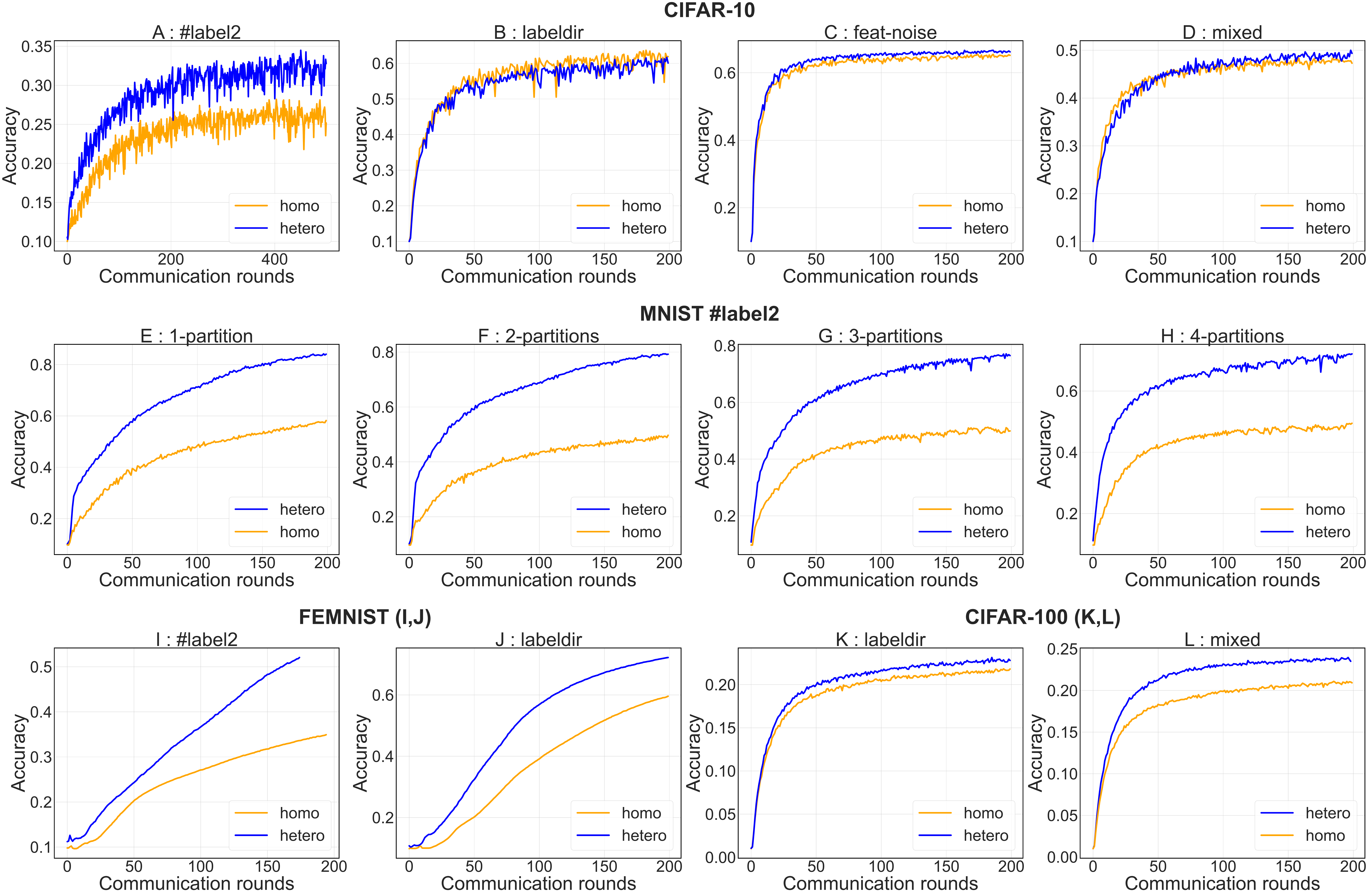} }}%
    \caption{Various top-1 test accuracy experiments conducted using MNIST, FEMNIST CIFAR-10 and CIFAR-100 as well as different models. Blue lines show locally heterogeneous graphs while orange lines show locally homogeneous graphs (details in Table \ref{tab:comparison1}, \ref{tab:comparison2}). }%
    \label{fig:plots}%
\end{figure*}

We mixed and matched datasets and models to better understand the effect of topology under a variety of scenarios. In our experiments, we used a 2-layer feed-forward model, a 2-layer CNN model and Resnet18 \cite{he2016deep}. We also utilized MNIST \cite{lecun-mnisthandwrittendigit-2010}, FEMNIST \cite{caldas2018leaf}, CIFAR-10 and CIFAR-100 \cite{krizhevsky2009learning} datasets for our training data. We used 1000 uniformly sampled images from FMNIST \cite{xiao2017fashion} as global data when conducting MNIST/FEMNIST experiments. Similarly, we used 1000 samples of Tinyimagenet \cite{Le2015TinyIV} as global data when working with CIFAR-10/CIFAR-100. In the case of quantity-based label skew, each node is randomly assigned $2$ classes while samples are equally distributed across the nodes. This is done to simulate a severe and strict quantity skew. In the case of distribution-based label skew, each party is allocated a proportion of the samples of each label according to Dirichlet distribution. I.e. Each node $v$ is allotted  $p_{k,v} \sim Dir_N (\beta = 0.5, 0.1)$ proportion of the instances of class $k$. Here, the concentration parameter $\beta$ determines the extent to which the labels are skewed (i.e. smaller $\beta$ produces a higher skew). For feature skew and mixed non-IID settings, we set the feature noise level $\eta=0.2, 0.3$ respectively. The noise levels specify the different levels of Gaussian noise added to the peer local datasets. Specifically, given noise level $\eta$, we apply noise  $\widetilde{x_i} \sim Gau(\eta \; . \; N_i/N)$, to each peer $N_i$ , where $Gau(\eta \; . \; N_i/N)$ is a Gaussian distribution with mean $0$ and variance $\eta \; . \; N_i/N$.

Furthermore, although $N$ clients participate in the morphing phase, we consider only a fraction of them as participants during phase III training. This is done for two reasons. First, some clusters in phase II, might have fewer nodes than others leading cliques to have uneven number of nodes. Hence, we try to even the cliques by attempting to sample $logN$ nodes from each cluster. This way, we create a fair comparison between heterogeneous and homogeneous graphs of comparable node sizes.  Second, by introducing a deficiency of peers (ergo a lack of training data), we further exacerbate the non-IID condition. In practice, a general lack of representative training data might occur especially when there is a relatively low number of active participants. Hence, it is interesting to examine the robustness of our approach under such highly skewed scenarios. Therefore, although the total global training data has been divided amongst $N$ nodes, we specify $N_p/N$ (in Table \ref{tab:comparison1}, \ref{tab:comparison2}) to denote that only $N_p$ nodes participated in phase III (training).


\subsection{Convergence Experiments}

We compare the average Top-1 accuracy of locally heterogeneous graphs (in blue) against locally homogeneous graphs (in yellow) of similar size and from the same distribution as shown in Fig. \ref{fig:plots}. In the case where we have few nodes (eg. experiments A,B,C,D,E,F,G,H,K,L), the competing topologies are composed of nodes that are almost entirely rearranged. I.e., despite having similar participants and communication overhead, the performance gains from a mere rearrangement of nodes can be quite significant. On the other hand, the scaled experiments I,J, show competing topologies composed of arbitrary nodes sampled after the clustering step in phase II. These experiments demonstrate how topological pre-processing can be used to select ideal participants that can increase the overall performance of the system. On the other hand, ignoring the pre-processing phases may result in a significantly lower convergence as shown by the locally homogeneous graphs.

As the CFIAR-10 experiments show, when fewer nodes are participating, locally heterogeneous graphs significantly improve quantity-based label skew without hurting performance under other non-IID settings. In the CIFAR-100 distribution-based label skew experiments, we observed that the smaller the concentration parameter $\beta$, the better locally heterogeneous graphs performed. This is because a smaller $\beta$ results in higher label skew making it easier to cluster in phase II. There is a small gain on feature noise and mixed non-IID cases. Overall, the small scale experiments show that topological pre-processing excels under quantity-based label skew. As we scaled the experiments using the FEMNIST dataset, the effects of local heterogeneity get more pronounced under the distribution-based label skew. As, the number of cliques/neighborhoods increases, so too does the influence of local heterogeneity. This shows topological pre-processing is ideal for scaling. 

\begin{table}[]
\centering
\caption{Robustness under different non-IID settings in Fig. \ref{fig:plots}. $\Delta$ Acc. shows the Top-1 test accuracy difference between heterogeneous and homogeneous lines.}
\label{tab:comparison1}
\resizebox{1\columnwidth}{!}{%

\begin{tabular}{ |ccccccc| }
\hline
\textbf{Experiment} &
  \textbf{\begin{tabular}[c]{@{}c@{}}Dataset\end{tabular}} &
  \textbf{\begin{tabular}[c]{@{}c@{}}Non-IID\end{tabular}} &
  \textbf{\begin{tabular}[c]{@{}c@{}}Model\end{tabular}} &
  \textbf{Nodes} &
  \textbf{\begin{tabular}[c]{@{}c@{}}$\Delta$ Acc.\end{tabular}} & \\ \hline
  
 A  & \multirow{4}{*}{CIFAR-10}  & \#label2 &  \multirow{4}{*}{ResNet-18} & 18/20 & \textbf{+6.28\%} & \\
 B                        &  &    labeldir  $\beta=0.5$ &   &        15/20                   &    -1.98\% &  \\
 C                        &  &     feat-noise $\eta=0.2$ &   & 20/24  & +1.13\% &   \\
 D                        &  &    mixed $\beta=0.5, \eta=0.3$ &   &        20/24           &       +1.59\%  & \\ \hline
 I &  \multirow{2}{*}{FEMNIST} & \#label2 &  \multirow{2}{*}{2-layer FFNN} & 122/1200 & \textbf{+18.4\%} &    \\ 
J                         &  &    labeldir $\beta=0.5$ &   &           118/1200                &       \textbf{+12.6\%} & \\ \hline
K &  \multirow{2}{*}{CIFAR-100} & labeldir $\beta=0.1$ &  \multirow{2}{*}{ResNet-18} & 22/24 & +1.31\% &    \\ 
L                         &  &    mixed $\beta=0.1, \eta=0.3$ &   &           22/24                &       +2.86\% & \\ \hline
\multicolumn{6}{l}{$^{\mathrm{a}}$Bold $\Delta$ Acc. values show values $\geq 5\%$}
\end{tabular}%

}
\end{table}

\begin{table}[]
\centering
\caption{Robustness under quantity-based label  skew and network partitions in Fig. \ref{fig:plots}. $\Delta$ Acc. shows the Top-1 test accuracy difference between heterogeneous and homogeneous lines.}
\label{tab:comparison2}
\resizebox{1\columnwidth}{!}{%
\begin{tabular}{ |ccccccc| }
\hline
\textbf{Experiment} &
  \textbf{\begin{tabular}[c]{@{}c@{}}Dataset\end{tabular}} &
  \textbf{\begin{tabular}[c]{@{}c@{}}Partition\end{tabular}} & 
  \textbf{\begin{tabular}[c]{@{}c@{}}Model\end{tabular}} &
  \textbf{Nodes} &
  \textbf{\begin{tabular}[c]{@{}c@{}}$\Delta$ Acc.\end{tabular}} & \\ \hline

 E  & \multirow{4}{*}{MNIST} &  1  & \multirow{4}{*}{2-layer CNN} & 22/24  & \textbf{+25.9\%} &  \\
 F                        &   &    2                     &  &        22/24                   &    \textbf{+29.6\%} & \\
 G                        &  &    3                    &   & 24/24  & \textbf{+25.8\%} &   \\
 H                        &   &    4                   &  &       24/24                   &      \textbf{+22.6\%} & \\ \hline
\end{tabular}%
}
\end{table}

\subsection{BFTM Analysis}

Assume we have a DL graph $G = <N,E>$ with a max neighbor size $M = \log(N)$, we initialize a pool of nodes $P$ with all the nodes in $N$. Given node $v \in N$, let $\beta^v_i$ be the total number of nodes encountered by node $v$ at the end of iteration $i$. $\beta^v_i = \sum_{j=0}^{i} \beta^v_j$. I.e. $\beta^v_i$ represents the number of nodes that have computed pairwise similarity with $v$ by iteration $i$. Let, $C$ represent node $v$'s cache. We initialize $C$ with $\O$. In every round, we add $v$'s 1-hop neighbors to $C$. 
In each iteration, node $v$ has $M$ 1-hop neighbors which all have $M - 1$ other 1-hop neighbors. Let's assume there are no connections among $v$'s 1-hop and 2-hop neighbors to maximize $v$'s encounter set. In each iteration, node $v$ downloads proxies from its $M$ 1-hop neighbors. Thus, node $v$ will directly compute its pairwise similarities against $M$ neighbors. Similarly, $v$'s $M$ 1-hop neighbors will download proxies from $v$ and $v$'s 2-hop neighbors to compute pairwise similarity computations. As such, they are in a position to compute node $v$'s pairwise similarity against its $M(M-1)$ 2-hop neighbors. Therefore, in one round, node $v$ will encounter $M + M(M-1) = M^2$ nodes. Finally, node $v$ will cache $M$ proxies from it's 1-hop neighbors as $v$'s $M$ 1-hop neighbors reciprocally cache node $v$'s proxy. Thus we work out node $v$'s per-round encounters ($\beta^v_i$) as follows:

\begin{equation*}
\begin{split}
\beta^v_i &=\sum_{k = 0}^{i} (M + M(M-1) + kM^{2}) \\    
&= \sum_{k = 0}^{i} M^2 + kM^2\\
&= M^2\sum_{k = 0}^{i}1 + k\\
&= M^2 ( i + \frac{i(i+1)}{2})\\
&= \frac{1}{2}i^2M^2 + \frac{3}{2}iM^2\\
\end{split}
\end{equation*}
To prove that $\beta^v_i = O(i^2M^2)$

\begin{equation*}
\begin{split}
\frac{1}{2}i^2M^2 + \frac{3}{2}iM^2 &\leq \frac{1}{2}i^2M^2 + \frac{3}{2}i^2M^2 \leq 2i^2M^2\\
\end{split}
\end{equation*}
Thus for $C = 2$, $i \geq 1$, $\beta^v_i = O(i^2M^2)$\\ To prove that $\beta^v_i = \Omega(i^2m^2)$
\begin{equation*}
\begin{split}
\frac{1}{2}i^2M^2 + \frac{3}{2}iM^2 &\geq \frac{1}{2}i^2M^2\\
\end{split}
\end{equation*}
Thus for $C = \frac{1}{2}$, $i \geq 1$ $\beta^v_i = \Omega(i^2M^2)$\\
Since, 
\begin{displaymath}
\frac{1}{2}i^2m^2 \leq \beta^v_i \leq 2i^2M^2
\end{displaymath}
Thus, by definition $\beta^v_i = \Theta(i^2M^2)$ for $C_1 = \frac{1}{2}, C_2 = 2, i \geq 1$. Finally, we want to show how many iterations it takes $v$ to encounter all $N-1$ neighbors. We solve for $\beta^v_i = N$.
\begin{equation*}
\begin{split}
\beta^v_i &= N \\
M^2i^2 &= N \\
\frac{1}{M^2} \Big(M^2i^2 \Big) &= \frac{1}{M^2}N \\
i^2 &= \frac{N}{M^2}\\
i &= \frac{\sqrt{N}}{logN} \\
\end{split}
\end{equation*}
Thus $i = \Theta(\frac{\sqrt{N}}{logN})$ $\blacksquare$

\subsection{Robustness Against Network Partitions}

To study the effect of network partitions on global convergence, we split the final graph into two, three and four disjoint sub-graphs immediately before phase (Phase III) as shown in Fig. \ref{fig:hetero-homo-ring}. The disjoint sub-graphs end up converging independently as updates no longer ripple between them. Our experiments in Fig.~(\ref{fig:plots} E,F,G,H) show locally heterogeneous sub-graphs maintain their convergence edge with respect to their locally homogeneous counterparts. This demonstrates the robustness of local heterogeneity in the presence of network partitions.

\subsection{Topological Pre-processing Overhead}

\begin{figure}
\begin{minipage}[t]{1.0\columnwidth}
  \includegraphics[width=\linewidth]{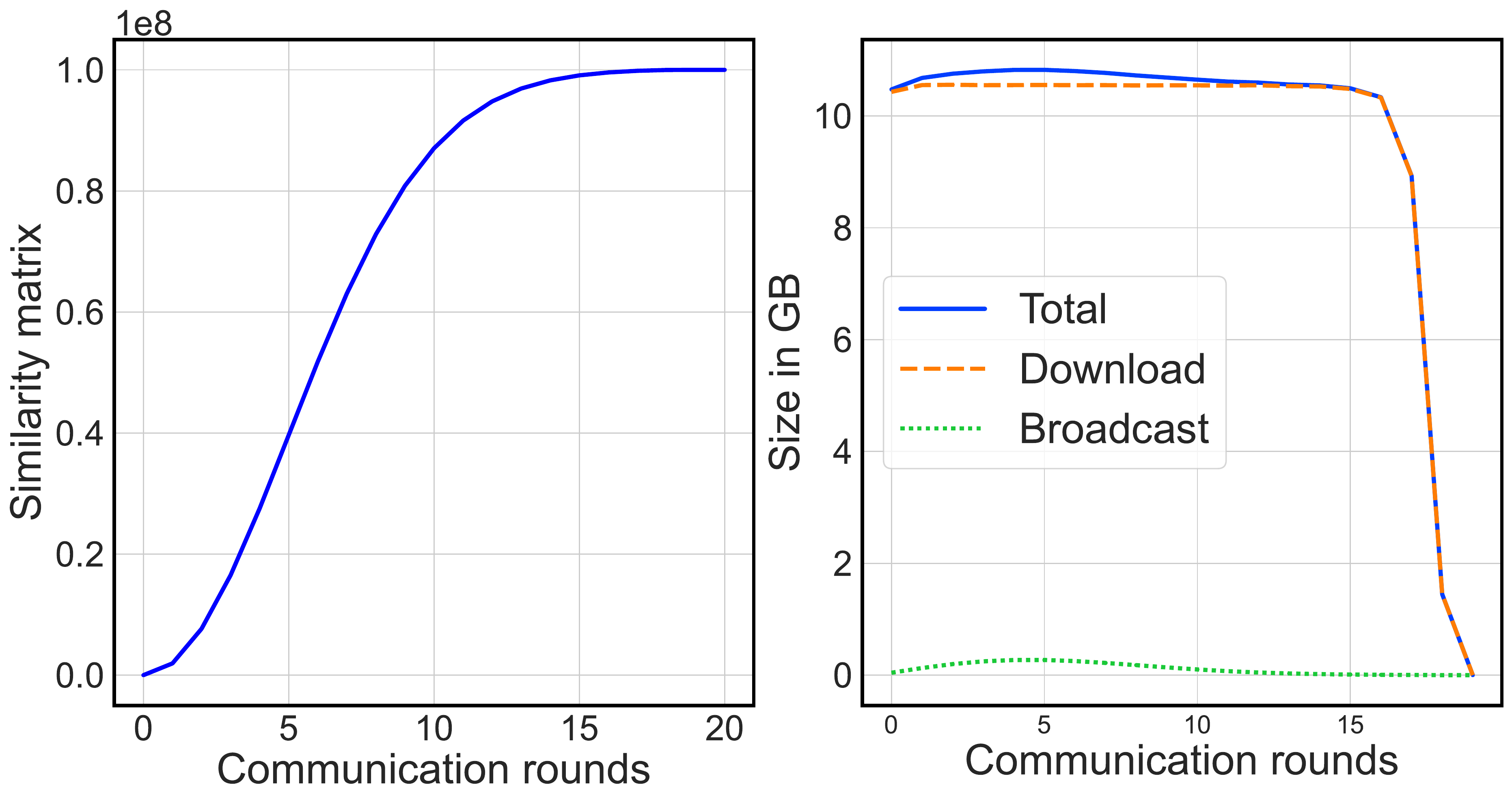}
\end{minipage}\hfill 
\caption{(Left) shows the similarity matrix population growth. (Right) shows the total communication size in gigabytes.}
\label{fig:10k-summary}%
\end{figure}

As described in the previous section, topological pre-processing incurs an initial cost. Here, we calculate the communication overhead required by phases I and II in a large DL system. Assume we have a DL system comprising $10^{4}$ nodes. Empirically, we're able to show that it requires $20$ BFTM rounds to conduct phase I (Fig. \ref{fig:10k-summary}). At the start of each round, nodes communicate proxies with a max of approximately $14$ other nodes. Furthermore, at the end of each round, nodes broadcast newly created similarity tuples. The broadcast bump in Fig. \ref{fig:10k-summary} (left) is proportional to the rise in matrix population shown in Fig. \ref{fig:10k-summary} (right). Total proxy download overhead is $191.5$GB ($\approx 20$MB per node). Total broadcast received by every node is about $2.235$GB. Thus total download per node equals $\approx 2.26$GB which is significantly small compared to the network size. In contrast, a naive proxy broadcast algorithm would require every node to download $762$GB worth of data.

To put it in perspective, in phase III, an image classification task using ResNet-18 and an image input size of 224x224 requires each node to download 14 models per round. I.e. each node will download $\approx 630$MB per communication round. Hence, topological pre-processing incurs an additional download overhead of about 4 communication rounds. As the task gets more difficult and the model size increases, the topological processing overhead remains constant. For instance, using a VGG16 model under the same experiment setting, will result every node to download $\approx 7.4$GB in each communication round of phase III. This means, the communication overhead incured by topological pre-processing is just a fraction of a single communication round.

\subsection{Ablation Study}

We further conducted experiments using gradient-based proxies as zero-data alternatives to KL-based proxies. We extracted last layer gradients from the last epoch of the prologue training to serve as model proxies. While this approach has the advantage of eliminating the need of a public dataset, our evaluation showed that it lacks stability under quantity-based label skew and suffered some convergence loss in the distribution-based label skew experiments. Moreover, it is difficult to use gradients of all layers for two reasons. First, it's expensive to communicate and second, such high dimensional vectors make it hard to derive singular (or fewer) descriptive values that represent the underlying models. Here, applying salient gradient selection and an effective similarity measure is a matter that requires further research. Another drawback of gradient proxies is the challenge that comes when dealing with varying model architectures. Under such circumstances, direct gradient comparison is not applicable.

\section{Conclusion}

In line with previous works, we demonstrate how locally heterogeneous DL graphs exhibit improved convergence compared to locally homogeneous graphs under non-IID settings. As such, we present a strong case for factoring in topology when designing a DL system. In our work, however, we measure heterogeneity in terms of model knowledge as opposed to underlying data distribution. To this end, we use proxies: soft-labels outputted on a global dataset as abstract representations for model knowledge. While earlier works require prior knowledge of local data distributions, we avoid peeking into local data, thus maintaining data privacy. We crafted a novel gossip algorithm called BFTM for efficiently computing pairwise proxy similarities using 1-hop communication and iterative graph restructuring. We then applied unsupervised clustering and uniform sampling to efficiently construct final candidate topologies with increased global convergence. Our experiments show higher yields in overall model accuracy under various non-IID settings as compared to locally homogeneous graphs with similar size and from the same distribution. Moreover, we demonstrate the scalability of the proposed pre-processing steps and the robustness of our topology under network partitions. Finally, we would like to point out that BFTM is agnostic towards the choice of proxy. As such, we can customize our similarity computation as needed. Furthermore, our topology based solution is orthogonal to other methods involving model compression \cite{Yu_2021_ICCV} and aggregation techniques \cite{yu2021spatl, yu2022resource} used to improve performance. Hence, our solution is quite adaptable to other DL experiments.

\section{Acknowledgement}

We would like to thank Emmanuel Mihret Abebe for his contribution in formalizing the BFTM time-complexity proof.

{\small
\bibliographystyle{ieee_fullname}
\bibliography{Paper}
}

\end{document}